\documentclass[journal,transmag]{IEEEtran}
\ifCLASSINFOpdf
  \usepackage[pdftex]{graphicx}
  \graphicspath{{../pdf/}{../jpeg/}}
  \DeclareGraphicsExtensions{.pdf,.jpeg,.png}
\else
  \usepackage[dvips]{graphicx}
  \graphicspath{{../eps/}}
  \DeclareGraphicsExtensions{.eps}
\fi
\usepackage{amssymb}
\usepackage{color}
\usepackage{amsmath}
\usepackage{booktabs}
\usepackage{multirow}
\usepackage[switch]{lineno}
\usepackage[colorlinks,linkcolor=green]{hyperref}
\begin{document}
%
\title{Robust Collaborative Learning of Patch-level and Image-level Annotations for Diabetic Retinopathy Grading from Fundus Image}


\author{\IEEEauthorblockN{Yehui Yang\IEEEauthorrefmark{1*},
Fangxin Shang\IEEEauthorrefmark{1*},
Binghong Wu\IEEEauthorrefmark{1}, 
Dalu Yang\IEEEauthorrefmark{1}, 
Lei Wang\IEEEauthorrefmark{1},
Yanwu Xu\IEEEauthorrefmark{1+}, \\
Wensheng Zhang \IEEEauthorrefmark{2} and
Tianzhu Zhang \IEEEauthorrefmark{3}
}
\IEEEauthorblockA{\IEEEauthorrefmark{1} Artificial Intelligence Group, Baidu, Beijing, China}
\IEEEauthorblockA{\IEEEauthorrefmark{2} Institute of Automation, Chinese Academy of Sciences, Beijing, China}
\IEEEauthorblockA{\IEEEauthorrefmark{3} University of Science and Technology of China, Hefei, Anhui, China}

\thanks{* Equal Contribution: yangyehuisw@126.com}
\thanks{+ Corresponding Author}}

%



\IEEEtitleabstractindextext{%
\begin{abstract}
Diabetic retinopathy (DR) grading from fundus images has attracted increasing interest in both academic and industrial communities.
Most convolutional neural network (CNN) based algorithms treat DR grading as a classification task via image-level annotations.
However, these algorithms have not fully explored the valuable information in the DR-related lesions. 
In this paper, we present a robust framework, which collaboratively utilizes patch-level and image-level annotations, for DR severity grading. By an end-to-end optimization, this framework can bi-directionally exchange the fine-grained lesion and image-level grade information. As a result, it exploits more discriminative features for DR grading. 
The proposed framework shows better performance than the recent state-of-the-art algorithms and three clinical ophthalmologists with over nine years of experience. 
By testing on datasets of different distributions (such as label and camera), we prove that our algorithm is robust when facing image quality and distribution variations that commonly exist in real-world practice. We inspect the proposed framework through extensive ablation studies to indicate the effectiveness and necessity of each motivation. The code and some valuable annotations are now publicly available.
\end{abstract}

\begin{IEEEkeywords}
Convolutional neural networks, Diabetic retinopathy, Fundus image, Collaborative learning
\end{IEEEkeywords}}

\maketitle

\IEEEdisplaynontitleabstractindextext

\begin{figure}[!t]
\centering
\includegraphics[width=3in]{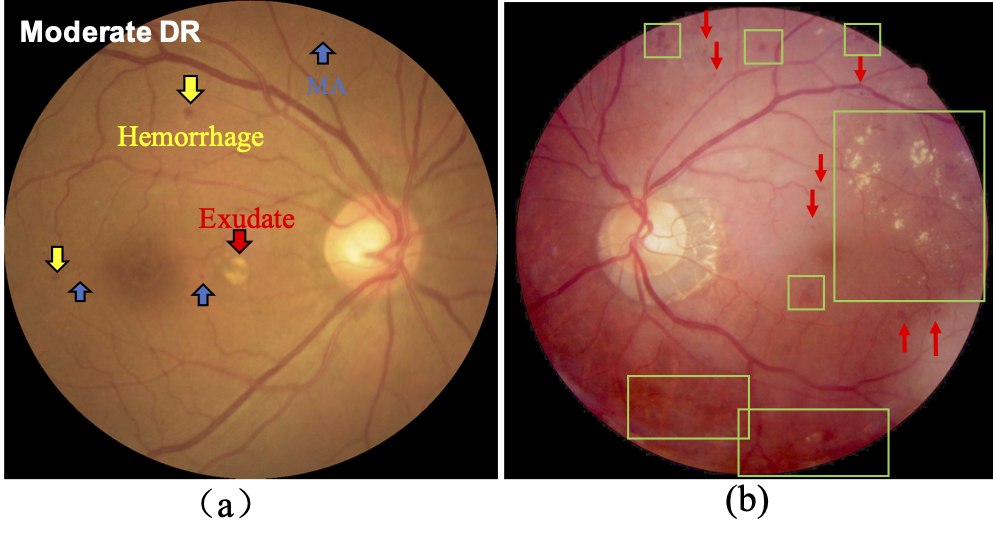}
\caption{a) A sample of fundus image with moderate DR, and the arrows indicate some key DR-related lesions. b) The green boxes are labeled by one of our annotators. The missing-annotated samples, indicated by the red arrows, may confuse the detectors that are trained with the entire images.}
\label{fig.fundusSample}       
\end{figure}


\section{Introduction}

Diabetes is a universal chronic disease that affects one in every eleven adults worldwide, and approximately $40\%$ to $45\%$ of patients with diabetes may develop diabetic retinopathy (DR) in their lifetime \cite{Gargeya2010aao,Haloi2016,Shaw2010,Pratt2016}.  According to \cite{googleJAMA,Kocur2002}, DR is one of the leading causes of irreversible blindness, while most parts of the world are short of qualified ophthalmologists. Therefore, the quick and automatic grading of the DR severity is critical and urgent to reduce the burdens of the limited ophthalmologists and provide timely morbidity diagnosis for massive patients.

DR grading aims to classify fundus images into different classes in terms of DR severity. According to the {\em International Clinical Diabetic Retinopathy Disease Severity Scale} \cite{DRstages}, DR falls into five severity grades including no DR, mild, moderate, severe, and proliferative. The five grades can also be merged as a binary classification, i.e. No-DR versus DR, or non-referable (no and mild DR) versus referable (moderate and worse DR) \cite{googleJAMA,Wang2017Zoom}.  
Recently, some researchers trend to leverage the powerful CNNs (convolutional neural networks) for DR grading. Researchers from Google Research use the Inception-v3 \cite{googleJAMA} to detect referable DR and macular edema. 
Sankar {\em et al.} \cite{Sankar2016}, Alban {\em et al.} \cite{stanford2016} and Pratt {\em et al.} \cite{Pratt2016} construct multi-class classifiers for DR grading with some popular or their own CNN architectures. 

However, the aforementioned end-to-end algorithms take the DR grading as a {\em black box} classification task, which ignores the valuable fine-grained DR-related lesions. According to \cite{DRstages}, the DR severity grades closely relate with the presence and absence of different lesion types. Fig. \ref{fig.fundusSample}(a) illustrates moderate DR and the corresponding lesion types of microaneurysms(MA), hemorrhages, and exudates.

Therefore, some researchers attempt to improve the grading performance by integrating the lesion information. Antal and Hajdu \cite{Antal2014} propose an ensemble-based algorithm for MA detection, mapping the fundus images into `DR/non-DR' based on the presence or absence of the MAs. Yang {\em et al.} \cite{myMiccai} introduce an offline lesion-based weighting scheme to improve the performance of DR grading. Lin {\em et al.} \cite{Lin2018Framework} propose a similar two-stage framework to integrate the patch-level lesion features and image-level global features with an attention network. 
Zhou {\em et al.} \cite{zhouCVPR2019} collaboratively optimize the lesion segmentation and DR grading in an adversarial way. 

Although the above works associate lesion information with DR grading, they still have the following issues:

1) Most of the above works construct a one-way feature transmission from lesion features to DR grades in a two-stage manner, i.e., the lesion-related modules and DR grading modules are trained separately without end-to-end learning. The lesion and grade features cannot be jointly fine-tuned for the final tasks. In this case, the lesion detectors need to be trained with large amounts of annotated data to provide a precise input for the following grading step; 2) Although \cite{zhouCVPR2019} can  optimize  lesion segmentation and DR grading modules in an end-to-end manner, they crave for pixel-level annotations to generate the lesion mask. Obviously, pixel-level annotations are extremely labor-consuming and expensive, especially medical annotations that require the dedication of domain experts. 

In this paper, we propose a robust end-to-end framework to collaboratively learn from both patch-level lesion and image-level grade annotations (CLPI) for DR grading. The proposed framework mainly consists of a lesion attention generator and a grading module. 
By training the lesion attention generator with only a few patch-level annotations, we can provide the grading module with patch-level attention of the input image in a semi-supervised manner. The grading module is to directly predict the DR severity grade based on the lesion attention and the input image. Additionally, our lesion attention generator can be pre-trained with image patches to avoid the missing label problem. As seen in Fig.~\ref{fig.fundusSample}, the missing labels are commonly existing in the annotation of medical images, which may confuse the detectors trained with the entire images \cite{xu2019missing}. The source code can be found at: {\em \href{https://github.com/PaddlePaddle/Research/tree/master/CV/CLPI-Collaborative-Learning-for-Diabetic-Retinopathy-Grading}{https://github.com/clpicode}}\footnote{The complete url: {\em https://github.com/PaddlePaddle/Research/tree/master/ \\
CV/CLPI-Collaborative-Learning-for-Diabetic-Retinopathy-Grading}}.

The main contributions of this paper can be highlighted as three-folds:
\begin{itemize}
\item[1)] We propose a robust collaborative learning framework to integrate the patch-level lesion and image-level grade annotations for DR grading. Experimental comparisons prove that CLPI has superior performance to the relevant state-of-the-art (SOTA) algorithms. Moreover, the proposed algorithm also achieves comparable performance of three clinical experienced ophthalmologists with over nine years of experience. By training and testing on the datasets from totally different distributions, the proposed CLPI presents robust performance compared with the alternative popular CNN classifiers.
\item[2)] We design a novel network architecture, i.e., lesion attention generator, which can generate the patch-level lesion attention map of an entire image with only one forward pass. Experiments prove that our lesion attention generator can effectively improve the performance of DR grading methods. This architecture can be trained with image patches instead of the entire images, which can alleviate missing label problems.
\item[3)] Extensive ablation studies have experimentally proved the contributions of lesion features for DR severity grade, as well as the necessity of building a bidirectional path to exchange information in an end-to-end manner between lesion module and grade module.
\end{itemize}

\begin{figure*}[!t]
\setlength{\abovecaptionskip}{-0.3cm}
\setlength{\belowcaptionskip}{-0.5cm}
\centering
\includegraphics[width=6.5in]{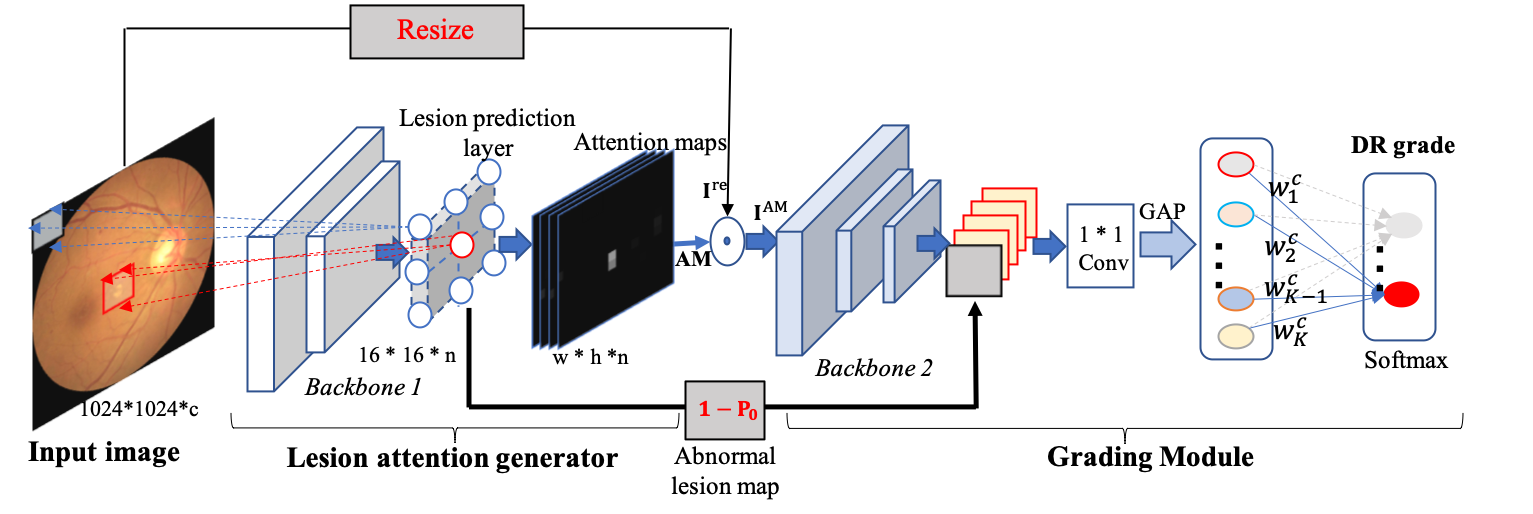}
\caption{Main workflow of the proposed algorithm. The architecture can be roughly dissected into 1) Lesion attention generator which consists of {\em Backbone 1} and attention map generation operations. 2) Grading module which consists of {\em Backbone 2} and the shortcut connection delivered abnormal lesion maps close to the classification head.}
\label{fig.mainWorkflow}       
\end{figure*}

\section{Related Work}

Automatic medical image analysis has attracted tremendous research interests because of its enormous demand, such as cancer detection based on computed tomography (CT) or whole sliding image (WSI) \cite{qi2020automated,lin2015inference,Wang2020Weakly}, fetal standard planes detection via ultrasound imaging \cite{Chen2017Ultrasound} and brain disease prognosis using magnetic resonance imaging (MRI). In this paper, we focus on the DR severity grading from fundus image. 
The early attempts for DR grading from the fundus images were usually done with two steps: handcrafted feature extraction and classification \cite{Pinz1998Mapping,Sopharak2009Automatic}. Acharya {\em et al.} \cite{Acharya2009Computer} used image-processing techniques to extract features from blood vessels and some key lesions and then classified the fundus images into five grades. Recently, CNN-based algorithms have shown superior performance in computer vision and brought powerful tools for DR assessment \cite{Alexnet,arcadu2019deep,nielsen2018deep,LiTMI2020}. Unlike handcrafted features, these deep networks can automatically learn discriminative features from large scale data.

Since DR severity grades have a high correlation with lesions existing in the fundus, some researchers also try to detect some key DR-related lesions \cite{Adal2018Automated,Playout2019}.
Pixel-level lesion segmentation is a popular way to accurately achieve both the locations and contours of the lesions. Eftekheri {\em et al.} \cite{Eftekheri2019Microaneurysm} aim to segment MAs out of fundus images via a two-step CNN. Chudzik {\em et al.}  \cite{Chudzik2018Exudate} utilize fully convolutional neural networks (FCNNs) for exudate segmentation. Yan {\em et al.} \cite{yan2019Learning} propose a mutually local-global algorithm for lesion segmentation based on U-Net \cite{unet}. Due to the shortage of pixel-level annotated data, the results of the lesion segmentation approaches are far from promising in practice.

Another way to locate abnormal regions is patch-level lesion detection. Patch-level annotations are relatively easy to obtain compared with pixel-level annotations. Silberman {\em et al.} \cite{Silberman2010Case} extract SIFT (scale invariant feature transform) features in the image patches and utilized SVM (support vector machine) to distinguish patches with exudates. 
Haloi {\em et al.} \cite{Haloi2016,Haloi2015} achieve promising performance in MA and exudate detection based on sliding windows and CNN classifiers. Van Grinsven {\em et al.} \cite{TMI2016} propose a selective sampling method for fast hemorrhage detection. Srivastava {\em et al.} \cite{Srivastava2017} achieve robust performance in finding MAs and hemorrhages based on multiple kernel learning methods.

\begin{table}[!htb]
\vspace{-0.5cm}
\centering
\caption {Comparisons among the recent related work on some key concepts. The last column only refers to the annotation level for lesion detection. Compared to patch-level annotation, the pixel-level is more time and labor consuming.}
\setlength{\tabcolsep}{3.2pt}
\begin{tabular}{lcccc}
\hline
      & \begin{tabular}[c]{@{}c@{}} Lesion features \\ for DR grading \end{tabular} & \begin{tabular}[c]{@{}c@{}} End-to-end \\ learning \end{tabular}   & \begin{tabular}[c]{@{}c@{}} Annotation \\ level\end{tabular} \\
\hline
\cite{Pratt2016,googleJAMA,Sankar2016,stanford2016}  & $\times$ & $\times$ & - & - \\
Yang {\em et al.}\cite{myMiccai}  & $\checkmark$ & $\times$ & patch \\
Wang {\em et al.}\cite{Wang2017Zoom}   & $\times$ & $\times$ & - \\
Lin {\em et al.}\cite{Lin2018Framework}   & $\checkmark$ & $\times$ & patch \\
Zhou {\em et al.}\cite{zhouCVPR2019}  & $\checkmark$ & $\checkmark$ & pixel \\
CLPI   & $\checkmark$ & $\checkmark$ & patch \\
\hline
\label{tab.relatedWork}
\end{tabular}
\end{table}

Most works take DR grading and lesion detection separately, and only a few approaches integrate both lesion and grade information for DR assessment \cite{Antal2014,myMiccai,Lin2018Framework,zhouCVPR2019}. Inspired by some recent attention-based methods which integrate local and global features \cite{Chen2020TCYB,Chen2020TCYB,Fu2020TCYB}, we convert lesion information into an attention map and collaboratively learn the patch-level and image-level features. 

Table \ref{tab.relatedWork} lists some recent work for DR grading on some key concepts include 1) whether lesion features are applied for DR grading, 2) whether end-to-end learning is used for collaboratively integrating the lesion and grading features, and 3) the annotation level (patch or pixel) of training samples for achieving lesion features.

According to Table~\ref{tab.relatedWork}, CLPI is the first end-to-end algorithm that integrates patch-level lesion and image-level grade annotations for DR grading.

\begin{figure}[htb]
\setlength{\abovecaptionskip}{-0.3cm}
\setlength{\belowcaptionskip}{-0.5cm}
\centering
\includegraphics[width=3.5in]{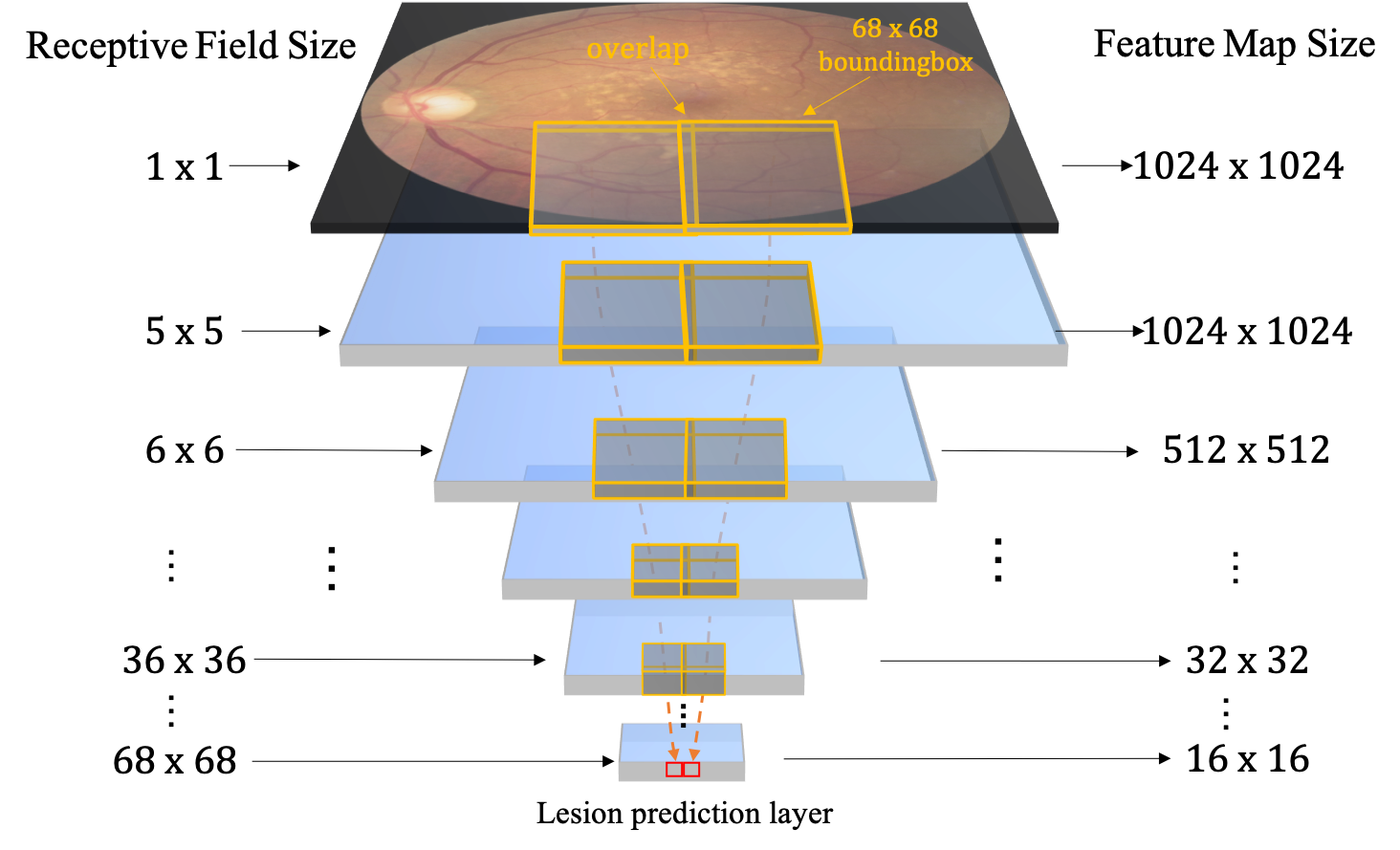}
\caption{Illustration of the receptive field design of the lesion attention generator.}
\label{fig.receptiveField}       
\end{figure}

\section{Methods}

In this section, we present the details of the proposed CLPI. 
As illustrated in Fig. \ref{fig.mainWorkflow}, the proposed algorithm can be dissected into the following parts:
1) A lesion attention generator to explore lesion features of an input image; 2) A grading module to classify the DR severity; 3) End-to-end learning details of feature integration for final decision.

\subsection{Lesion Attention Generator}

The lesion attention generator aims to explore lesion-related attention maps w.r.t. the input images. Firstly, an input image is split into patches, then a probability vector is generated for each patch according to the lesions existed in the patch. In this paper, the four dimensions of the probability vector correspond to the probabilities of MA, hemorrhage, exudate, or {\em none of the three lesions}, respectively. Ideally, a bigger value in the probability vector indicates a higher probability that the corresponding lesion exists in the input patch. Finally, the attention map of the entire image is constructed by extending the probabilistic vectors. 

A direct way to extract lesion information from an image is to apply some SOTA detectors (e.g. Faster-RCNN \cite{FasterRCNN}) for lesion detection. Another way is to take the lesion detection as a patch-based classification problem \cite{myMiccai}, where the patches are obtained by sliding windows. However, both the above approaches should be trained offline and cannot be embedded in an end-to-end learning framework to fine-tune the lesion attention for adapting the final DR grading.

Our proposed method can split the input image and generate an attention map with one forward pass. Accordingly, the lesion attention generator consists of two parts: a lesion detection backbone and an attention map generator. The details are as following:

\noindent {\bf Lesion Attention Generator Backbone:} To detect patch-based lesion information within one forward pass, we design the architecture of the detection backbone as Table \ref{tab.lesionNetworkBackbone}. The activation function between two convolution layers is the recently proposed Mish activation \cite{misra2019mish}, and the batch normalization (BN) \cite{BN} is applied before each Mish. We put a Softmax layer named the {\em lesion prediction layer} after the last convolution layer. The last column of Table \ref{tab.lesionNetworkBackbone} records the size of the feature maps in each layer when the size of the input image is $1024 \times 1024$ and $68 \times 68$, respectively.

The motivation of this architecture is to control the receptive field so that each unit in {\em lesion prediction layer} only covers a $68 \times 68$ region in an input image of size $1024 \times 1024$ (see Fig.~\ref{fig.receptiveField}).
As illustrated in Table~\ref{tab.lesionNetworkBackbone}, if the width and height of an input image are $1024 \times 1024$, the feature map of the {\em lesion prediction layer} is $\mathbf{P} \in \mathbb{R}^{16 \times 16 \times n}$, wherein each channel $\mathbf{P}_{i} \in \mathbb{R}^{16 \times 16} (i = 0,1,...,n-1)$ denotes the probabilistic matrix w.r.t the $i$-th class. In this paper, $n$ is equal to $4$ , indicating $3$ target lesions plus {\em no target lesion} (class $0$).
By this network design, a single forward pass can obtain the four-dimension probabilistic vectors of $16 \times 16$ patches in spatial order. This is equivalent to splitting the input image via a $68 \times 68$ sliding window with the stride $64$ then performing forward pass individually \footnote{If we control the receptive fields of prediction units as $64 \times 64$ non-overlapped regions, some lesions around the region boundaries may be missed by the detectors. Therefore, the receptive fields are designed with slightly overlapped in the lesion detection backbone.}. 

Moreover, the lesion attention generator can be trained via image patches instead of leveraging an entire image with full lesion annotations. According to the network architecture in Table \ref{tab.lesionNetworkBackbone}, a $1024 \times 1024$ image will generate a $16 \times 16$ lesion predictions. Wherein, each prediction corresponding to a $68 \times 68$ patch in the input image. However, if the input image is a $68 \times 68$ patch, the lesion network backbone will exactly generate a $1 \times 1$ prediction. This leads to an exciting advantage that our lesion attention generator trained with patches is equivalent to one trained with entire images, which can avoid the confusion brought by the underlying missing lesion labels (see the experimental evaluation in Section \ref{sec.discussion}). 

\begin{figure}[!htb]
\setlength{\abovecaptionskip}{-0.3cm}
\setlength{\belowcaptionskip}{-0.5cm}
\centering
\includegraphics[width=3in]{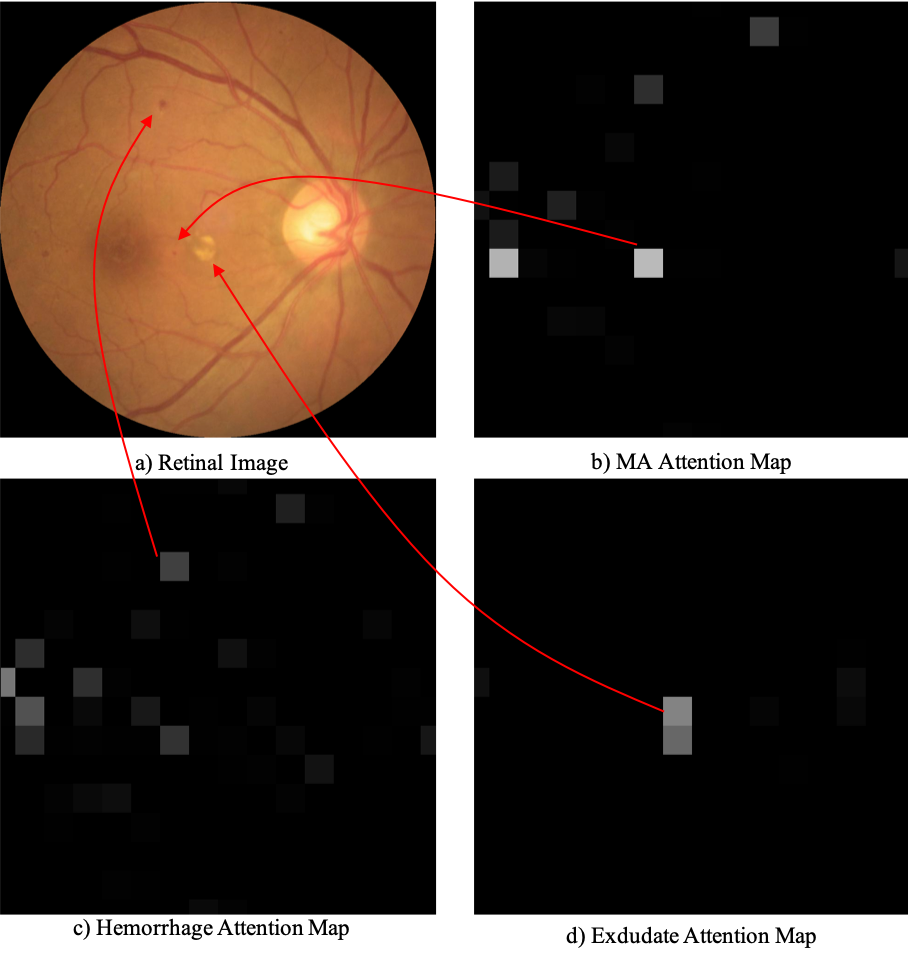}
\caption{ a) The fundus image with representative lesions; b), c) and d) are lesion attention maps corresponding to MA, hemorrhage and exudate respectively. We can see that the lesion regions will get relatively bigger attention values.}
\label{fig.lesionMap}       
\end{figure}

\noindent {\bf Attention Map Generation:} The attention map $\mathbf{AM} \in \mathbb{R}^{W \times H \times n}$ is constructed by expanding the lesion probabilistic matrix $\mathbf{P}$, where $W$ and $H$ denote the input width and height of the following severity grading net, respectively. In this paper, both $W$ and $H$ are set as $512$. Each entry of the probability matrix $\mathbf{P}_{i}$ is expanded to a $\frac{W}{16} \times \frac{H}{16}$ sub-matrix by duplicating, then each $\mathbf{P}_{i}$ will generate $16 \times 16$ sub-matrices. $\mathbf{AM}$ is constructed by tiling these sub-matrices together in spatial order. An example of the attention map generation procedure can be seen in the Appendix \ref{appendix.attentionMapGen}. As shown in Fig. \ref{fig.lesionMap}, the highlights convey the lesion information to construct the imbalanced attention map.

\begin{table}[!htb]
\vspace{-0.5cm}
\centering
\scriptsize
\caption{The Backbone Architecture of Lesion Attention Generator}
\label{tab.lesionNetworkBackbone}
\setlength{\tabcolsep}{3pt}
\begin{tabular}{c|c|c|c|c|c|c}
\hline
Layer & Kernel Size  & Stride & Padding & \begin{tabular}[c]{@{}c@{}}\textbf{Receptive field}\\ \textbf{of each unit}\end{tabular} &   \multicolumn{2}{c}{ \begin{tabular}[c]{@{}c@{}} $w \times h$ of the \\ feature map size \end{tabular}}  \\
\hline
Input & - & - & - & - & $1024 \times 1024$  & $68 \times 68$\\
\hline
Conv & 5 $\times$ 5 $\times$ 32 & 1 & 2 & 5$\times$5 &  $1024 \times 1024$  & $68 \times 68$\\
\hline
Conv & 2 $\times$ 2 $\times$ 64 & 2 & 0& 6 $\times$ 6 & $512 \times 512$ & $34 \times 34$\\
\hline
Conv & 2 $\times$ 2 $\times$ 128 & 2 & 0 & 8 $\times$ 8 & $256 \times 256$ & $17 \times 17$ \\
\hline
Conv & 2 $\times$ 2 $\times$ 256 & 2 & 0& 12 $\times$ 12 & $128 \times 128$ & $8 \times 8$\\
\hline
Conv & $2 \times 2 \times 512$ & 2 & 0 & 20 $\times$ 20 & $64 \times 64$ & $4 \times 4$ \\
\hline
Conv & 2 $\times$ 2 $\times$ 1024 & 2 & 0 & 36 $\times$ 36 & $32 \times 32$ & $2 \times 2$\\
\hline
Conv & 2 $\times$ 2 $\times$ 1024 & 2 & 0 & 68 $\times$ 68  & $16 \times 16$ &  $1 \times 1$\\
\hline
Conv & 1 $\times$ 1 $\times$ 1024 & 1 & 0& 68 $\times$ 68   & $16 \times 16$ &  $1 \times 1$\\
\hline
Conv & 1 $\times$ 1 $\times$ 4 & 1 & 0& 68 $\times$ 68  & $16 \times 16 $ &  $1 \times 1$\\
\hline 
Softmax & - & - & - & 68 $\times$ 68 & $16 \times 16$ &  $1 \times 1$\\
\hline
\end{tabular}
\vspace{-0.5cm}
\end{table}

\subsection{Grading Module} \label{subsec.globalNetwork}

The grading module, which contains the neck and head of the CLPI framework, is responsible for grading the severity of DR. In this paper, we set up two types of grading tasks: one is five-grade classification according to the {\em International
Clinical Diabetic Retinopathy scale} \cite{DRstages}, and the other is a binary classification task which wraps the above five grades into non-referable versus referable. 

As illustrated in Fig. \ref{fig.mainWorkflow}, there exist four key parts in the grading module: 1) A classification backbone without fully connected layer ({\em Backbone 2}); 2) A shortcut connection to deliver abnormal lesion map close to grading head; 3) A $1 \times 1$ convolutional layer to integrate the abnormal lesion map and the features from the classification backbone; 4) The global average pooling (GAP) and softmax layers for predicting the DR grades. 

In this paper, we use DenseNet-121 \cite{densenet} as the classification backbone by removing the fully connected layers. The abnormal lesion map transmitted by the shortcut is $\mathbf{1} - \mathbf{P}_0$, wherein $\mathbf{P}_0$ is the probability matrix of class $0$ (no target lesion) from the lesion prediction layer. As shown in Fig. \ref{fig.mainWorkflow}, we stack the abnormal attention map with the feature maps of the classification backbone to collaboratively learn both detection and classification nets.

The motivations of the abnormal shortcut between the {\em lesion prediction layer} and the output of the classification backbone are: (1) to directly back-propagate gradient to the lesion attention generator to avoid the potential gradient-vanishing; (2) to provide the semi-supervised lesion-related information closer to the prediction layer to improve the DR grading performance.

\subsection{End-to-end Collaborative Learning of CLPI Framework}

As shown in Fig.~\ref{fig.mainWorkflow}, a weighted attention map  $\mathbf{I}^{AM} \in \mathbb{R} ^{W \times H \times n*c}$  is constructed to feed the grading module. Let $\mathbf{I}^{re} \in \mathbb{R} ^{W \times H \times c}$ denotes the resized input image ($c$ is the channel of the input images). The weighted attention map can be calculated as

\begin{equation}
\begin{split}
\mathbf{I}^{AM}_{k} &= (\mathbf{AM}_{i} + \mathbf{1}) \odot \mathbf{I}^{re}_{j}, \\
s.t., k &= 0, 1, ..., n*c - 1, \\
i &= 0, 1, ..., n - 1, \\
j &= 0, 1, ..., c - 1,
\end{split}
\label{eq.MergeImLesion}
\end{equation}

where $\mathbf{I}^{AM}_{k}$ denotes the $k$-th channel of $\mathbf{I}^{AM}$. $\mathbf{AM}_{i}$ is the $i$-th channel of the attention map, and $\mathbf{I}^{re}_j$ denotes the $j$-th channel of $\mathbf{I}^{re}$. $\odot$ means the element-wise product. $\mathbf{1}$ is a constant matrix with the same size of $\mathbf{AM}_{i}$. It is used to prevent information dropout when the probabilities in $\mathbf{I}^{AM}_{k}$ are small.

Since the entries in the attention map $\mathbf{AM}_{i}$ implicitly carry the probabilities of having the $i$-th lesion in the patches, an image patch has a specific lesion with a higher probability will get a higher weight in the element-wise production. 
As seen in Fig.~\ref{fig.lesionMap}, imbalanced attentions are covered on the input image by highlighting the lesion patches.

To train the CLPI framework, first, we pre-train the lesion attention generator with patches. Then the entire CLPI framework is put into an end-to-end training by using only image-level DR grade labels. The classification backbone of the grading module can also be pre-trained for speeding up the convergence.

The loss function for both pre-training of lesion attention generator and the end-to-end training of the entire framework is the {\em cross-entropy loss}:
\begin{equation}
    CE(x_{i}, y_{i}, S) = - \sum_{c=1}^{M}\mathbf{1}(y_i=c)log(S(c|x_i))),
\end{equation}
where $x_i$ and $y_{i}$ denote the $i$-th input image/patch and the corresponding label respectively, and $S$ is the softmax output of the CNNs classifiers. $M$ is the class number, i.e. , $M = 4$ for the lesion attention generator. M = 2 and 5 for binary and multiple DR grading respectively. $\mathbf{1}(.)$ is an indicator function that is equal to $1$ when $y_i = c$, and $S(c|x_i))$ denotes the output probability of the unit w.r.t. class $c$.

\section{Experimental Results}

\subsection{Datasets}\label{sec.dataset}

\noindent {\bf Lesion Dataset:}  We use the public available {\em IDRiD Dataset} \cite{IDRiD} that provides $81$ fundus images ($54$ for training and $27$ for testing) with pixel-level annotations of lesions including MAs, hemorrhages and exudates. Since we only need patch-level lesion annotations, we turn the pixel-level annotation masks into $68 \times 68$ bounding boxes, and the label of each lesion patch is determined by the lesion type located in the center of this patch. If different lesion regions happen to share the same center, the corresponding patch will be given the label of the most severe lesion. Our severity rank is exudate $>$ hemorrhage $>$ MA. 
The distribution of the lesion patches is presented in Table~\ref{tab.idrid}.

\begin{table}[!htb] 
\vspace{-0.5cm} 
\centering
\caption {Distribution of IDRiD dataset. The last three columns indicate patch numbers}
\label{tab.idrid}
\setlength{\tabcolsep}{2.8pt}
\begin{tabular}{lc|ccc} 
\hline
 & Image number & MA  & Hemorrhage & Exudate   \\
\hline
Training & 57 & 2180 & 1431 & 6305 \\
Testing & 24 & 986 & 623 & 3474 \\
\hline
\end{tabular}
\vspace{-0.2cm} 
\end{table}

\noindent {\bf DR grade Dataset:} In this paper, the image-level datasets are listed as follows:
\begin{itemize}
    \item {\em Messidor-1 Dataset} \cite{messidor} contains $1,200$ fundus images from three French hospitals. However, their severity grade only has four levels, which is slightly different from the five-level international standard \cite{DRstages}. 
    
    \item {\em Messidor-2 Dataset} is an extension of the original Messidor-1 dataset, which contains $1748$ eye fundus images, and each image is classified into one of the five DR grades according to \cite{DRstages}.
    
    \item {\em LIQ-EyePACs} is a subset of the EyePACS dataset \cite{EyePACS} for evaluating the robustness of the grading methods, which contains some low image quality samples. Since there are a fair number of label biases in the original EyePACS dataset according to our cooperative ophthalmologists, we turn to {\em LIQ-EyePACs} under the following guidelines: 1) The labels of {\em LIQ-EyePACs} are rechecked by our cooperative ophthalmologists. 2) The quality of images is relatively low which contains noises like under/over-exposure and out-of-focus problems. These noises are commonly encountered in real-world practice. The detailed label distribution is in Table~\ref{tab.privateGradeSta}.

    \item {\em Private Datasets}: Our fundus images are collected from over $20$ hospitals. With the help of more than $30$ licensed ophthalmologists, the {\em private}  DR grade dataset has $36,270$ samples that contain one of five DR severity grade labels. To improve the overall annotation efficiency, each fundus image is first classified into the referable or non-referable DR by at least three licensed ophthalmologists, the binary groundtruths of the images are given by the majority voting. Based on the binary DR grade, three well-trained annotators or one licensed ophthalmologist will label the images with one of the five fine-grained DR grades according to \cite{DRstages}. 
\end{itemize}

The patch-level annotations of {\em IDRiD} and  the name list of {\em LIQ-EyePACs} will be provided upon request. 

\begin{table}[!htb]
\vspace{-0.5cm}
\centering
\caption {Distribution of {\em LIQ-EyePACs} and {\em Private} DR grade dataset.}
\label{tab.privateGradeSta}
\begin{tabular}{l|ccccc}
\hline
     & None DR & Mild & Moderate  & Severe  & PDR \\
\hline 
{\em LIQ-EyePACS} & 7,286 & 675 & 1,507 & 247 & 285 \\
\hline
{\em Private} & 19,826	& 3,220 & 9,760 &	2,069 & 1,395 \\
\hline
\end{tabular}
\vspace{-0.8cm} 
\end{table}

\subsection{Implementation details}

\noindent {\bf Data Preprocessing and Augmentation:} We subtract all the images by the local average color to highlight effective details on the fundus image, and the readers can turn to the competition report from the champion of the Kaggle competition \footnote{https://github.com/btgraham/SparseConvNet/blob/kaggle\_Diabetic\_\\ 
Retinopathy\_competition/competitionreport.pdf} for more details. We also apply some commonly used data augmentation strategies for both patches and the entire images, including randomly crop with scale=(0.9, 1.1), ratio=(0.9, 1.1), random horizontal and vertical flip (p=0.5), and random rotation with the rotate degree range in (0, 180). 

\noindent {\bf Training details:} 
The lesion attention generator is pre-trained with the patches from {\em IDRid dataset}. 
To train the final framework in an end-to-end manner, the gradients are {\bf only} back-propagated from the {\em abnormal} shortcut to fine-tune the lesion attention generator. The grade information from the shortcut can be more directly transmitted through the shortcut than from the {\bf deep} grading net. 

The training procedure consists of $100$ epochs with four Nvidia P40 GPU, and the batch size is $32$. To tune the learning rate, we utilized warmup strategy from 0 to 5e-2 in the first 5 epochs, then applied cosine decay in the remaining epochs. The code of the proposed algorithm is publicly available with the {\em PaddlePaddle deep learning platform}\footnote{https://github.com/PaddlePaddle/Paddle}. Given a $1024 \times 1024$ image, the inference time of our entire framework is 0.028 seconds per image with a single P40 GPU card. 

\noindent {\bf Evaluation Metrics:} For multi-grades classification, we use {\em Cohen Kappa values} which is a commonly used metric to measure the agreement between the predictions and the reference grades. Kappa values vary between $0$ (random agreement between raters) and $1$ (complete agreement between raters) \cite{cohenKappa}.  For the binary classification, we evaluate with AUC (area under the ROC curve) metric.

\begin{table*}[!htb] 
\vspace{-0.5cm}
\centering
\caption {Comparison of CLPI with the SOTA algorithms in {\em Messidor-1} dataset.}
\label{tab.messidorCmp}
\begin{tabular}{c|c|c|c|c|c|c} 
\hline
  &\begin{tabular}[c]{@{}c@{}}VNXK \\ \cite{vo2016new}\end{tabular} & \begin{tabular}[c]{@{}c@{}} CKML \\ \cite{vo2016new}\end{tabular} & \begin{tabular}[c]{@{}c@{}}Zoom-in-Net \\ \cite{Wang2017Zoom}\end{tabular}   & \begin{tabular}[c]{@{}c@{}}AFN \\ \cite{Lin2018Framework}\end{tabular}  & \begin{tabular}[c]{@{}c@{}}Semi+Adv \\ \cite{zhouCVPR2019}\end{tabular}    & CLPI \\
\hline
AUC of Non-referable / Referable DR  & 0.887 & 0.891 & 0.957 & 0.968 & 0.976 & \bf{0.985} \\
AUC of No-DR / DR & 0.870 & 0.862  & 0.921 & - & 0.943 & \bf{0.959} \\
\hline
\end{tabular}
\end{table*}

\subsection{The Effectiveness and Robustness of CLPI in DR Grading}

\noindent {\bf Comparison with SOTA algorithms:} Table \ref{tab.messidorCmp} lists the comparison between CLPI and some recent SOTA papers, including VNXK (VGGNet with extra kernel) \cite{vo2016new}, CKML (CNN-combined kernels with multiple losses network)\cite{vo2016new} Zoom-in-Net \cite{Wang2017Zoom}, AFN (Attention fusion network)\cite{Lin2018Framework} and Semi+Adv (semi-supervised and adversarial architecture) \cite{zhouCVPR2019}. VNXK and CKML are similar methods with different kernel strategies. Zoom-in-Net explores the suspicious regions of the fundus images in an unsupervised way for DR grading. AFN transmits lesion attention for DR grading in an offline way. Semi+Adv utilizes pixel-level lesion information to improve DR grading performance. All these algorithms are under the same evaluation protocols as presented in Zoom-in-Net and Semi+Adv. 
As seen in Table \ref{tab.messidorCmp}, our CLPI achieves superior performance to the SOTA algorithms in most cases. Some possible explanations include 1) The frameworks of VNXK, CKML and Zoom-in-Net not take the lesion annotations into consideration. 2) AFN learns an attention network which relies on both the lesion features and the entire images as parallel input. Therefore, AFN may not easily achieve favorable performance when the numbers of the lesion and grade annotations are imbalanced. The proposed lesion attention is initiated by the lesion annotation and then fine-tuned by the image-level annotations, which can alleviate the imbalanced issues between the lesion and grade annotations.
Moreover, trained with the same lesion and grade datasets in {\em IDRiD} and {\em Messidor-1}, the CLPI achieves comparable performance with \cite{zhouCVPR2019} that utilizes pixel-level lesion annotations. 

\noindent {\bf Comparison with Senior Ophthalmologists:} 

\begin{figure}[!htb]
\setlength{\abovecaptionskip}{-0.3cm}
\setlength{\belowcaptionskip}{-0.9cm}
\centering
\includegraphics[width=3in]{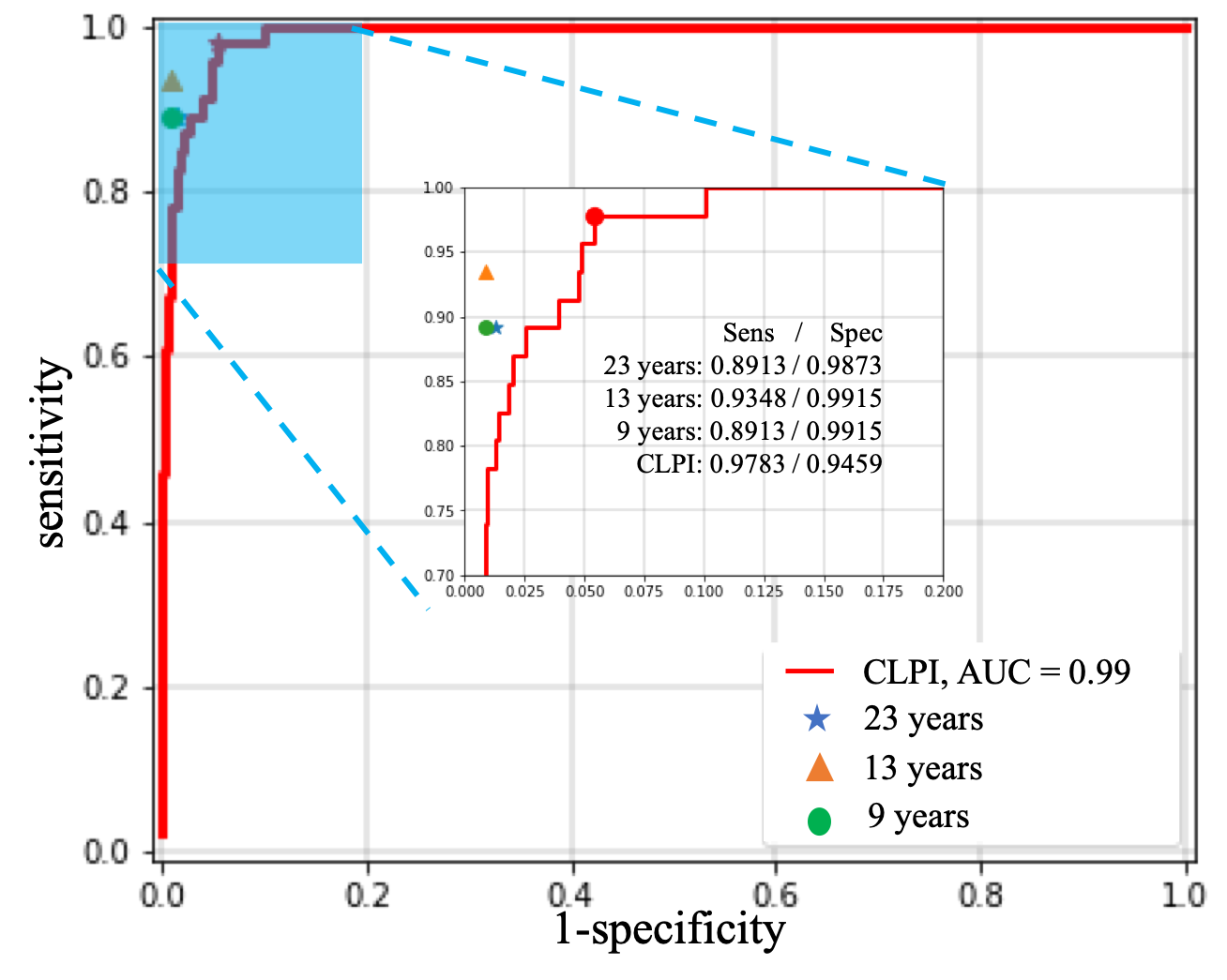}
\caption{The comparison between CLPI and three senior licence ophthalmologists on the same testing set. CLPI achieves comparable performance with the three over 9-years clinical experienced ophthalmologists.}
\label{fig.humanCmp}       
\end{figure}

In this section, we select the common annotations ($989$ fundus images from our {\em Private} dataset) of three senior ophthalmologists as DR testing set. Each ophthalmologists has a clinic experience of $23$, $13$ and $9$ years, respectively. All the three ophthalmologists label the testing images as referable or non-referable DR, and the groundtruths are achieved by majority voting. The sensitivities and specificities of the three ophthalmologists are shown in Fig.~\ref{fig.humanCmp} as well as the ROC (receiver operating characteristic) curve of CLPI. We can see that CLPI achieves comparable performance with the senior ophthalmologists in detecting referable DR.

\noindent {\bf The robustness of CLPI Against Commonly Existed Challenges In Real-world Practice:}

In real-world practice, compared with the training set, the testing images may be generated from totally different camera brands and label distributions. Additionally, the quality of the testing images can vary with different photographers and shooting environments. These issues bring inevitable challenges to the robustness of all the DR grading methods in reality.

To evaluate the robustness of different approaches, Table~\ref{tab.grading_cmp} lists the results of the models trained with one dataset while tested on another dataset with a totally different distribution and collected environment. {\em LIQ-EyePACs} is the dataset with low quality images that contains various noises including illumination, blur, artifacts problems.
All the popular CNNs architectures in Table~\ref{tab.grading_cmp}, including DenseNet121\cite{densenet}, ResNet50 \cite{ResNet} and Inception-V4 \cite{inceptionV4})),  are pre-trained with the same samples as CLPI for a fair comparison.

In the first several rows in Table~\ref{tab.grading_cmp}, our private dataset is randomly split into training, validation and testing sets by approximate $6:2:2$. All the models are only trained with our private training dataset and tested on the other datasets. Similarly, {\em Messidor-2} is split into training, validation and testing sets by $6:2:2$, and the last five rows in Table~\ref{tab.grading_cmp} are the results of the models that only trained with {\em Messidor-2} datasets.

We can see that the performance of all the methods decreases in the testing sets from different distributions. Meanwhile, the CLPI outperforms the alternative algorithms by a large margin in these cases, which proves the robustness of CLPI in facing different practical circumstances. 

In {\em Messdior-1} dataset, each image is annotated into one of four DR severity grades, which is different from the annotation standard compared to the datasets in Table~\ref{tab.grading_cmp}. Therefore, the results on {\em Messidor-1} are not recorded in the table.

\begin{table*}[!htb]
\caption {To evaluate the robustness of the algorithms, all the models are trained with one dataset and tested on the images from totally different distributions.}
\centering
\footnotesize
\label{tab.grading_cmp}
\begin{tabular}{l|c|cc|cc|cc}
\hline
\multirow{2}{*}{\begin{tabular}[c]{@{}c@{}}Training \\ sets \end{tabular}} & Testing sets  & \multicolumn{2}{c|}{\em Private} & \multicolumn{2}{c|}{\em Messidor-2} & \multicolumn{2}{c}{\em LIQ-EyePACS}\\
\cline{2-8}
&Methods & \begin{tabular}[c]{@{}c@{}}Five-grade\\ (Kappa)\end{tabular} & \begin{tabular}[c]{@{}c@{}}Binary\\ (AUC)\end{tabular} & \begin{tabular}[c]{@{}c@{}}Five-grade\\ (Kappa)\end{tabular} & \begin{tabular}[c]{@{}c@{}}Binary\\ (AUC)\end{tabular} & \begin{tabular}[c]{@{}c@{}}Five-grade\\ (Kappa)\end{tabular} & \begin{tabular}[c]{@{}c@{}}Binary\\ (AUC)\end{tabular}\\
\hline
\multirow{4}{*}{\em Private} 
&ResNet-50           & 0.885  & 0.980 & 0.651  & 0.874 & 0.629 & 0.844                 \\
&DenseNet-121        & 0.886  & 0.979 & 0.645  & 0.880 & 0.614 & 0.848               \\
&Inception-V4        & 0.899  & 0.980  & 0.667  & 0.887 & 0.625 & 0.849             \\
\cline{2-8}
& \textbf{CLPI} & \textbf{0.908} & \textbf{0.983} & \textbf{0.703} & \textbf{0.946} & \textbf{0.692} & \textbf{0.916} \\
\hline
\multirow{4}{*}{\em Messidor-2} 
&ResNet-50           & 0.682  & 0.916 & 0.793  & 0.945 & 0.492 & 0.840                 \\
&DenseNet-121        & 0.565  & 0.923 & 0.794  & 0.945 & 0.396 & 0.826               \\
&Inception-V4        & 0.497  & 0.811  & 0.803  & 0.954 & 0.312 & 0.685             \\
\cline{2-8}
&\textbf{CLPI} & \textbf{0.838} & \textbf{0.969} & \textbf{0.832} & \textbf{0.975} & \textbf{0.514} & \textbf{0.877} \\
\hline
\end{tabular}
\end{table*}

\subsection{Ablation Studies}

In this section, we experimentally investigate some key concerns of CLPI from the following aspects: 1) Our lesion attention generator is effective in exploring valuable lesion information for improving the DR grading performance. 2) It is necessary to optimize CLPI in an end-to-end manner, which can collaboratively  build bidirectional information exchange between image-level grade and five-grained lesion features, to achieve more promising results.

\begin{figure}[!htb]
\setlength{\abovecaptionskip}{-0.2cm}
\setlength{\belowcaptionskip}{-0.5cm}
\vspace{-0.5cm}
\centering
\includegraphics[width=3.6in]{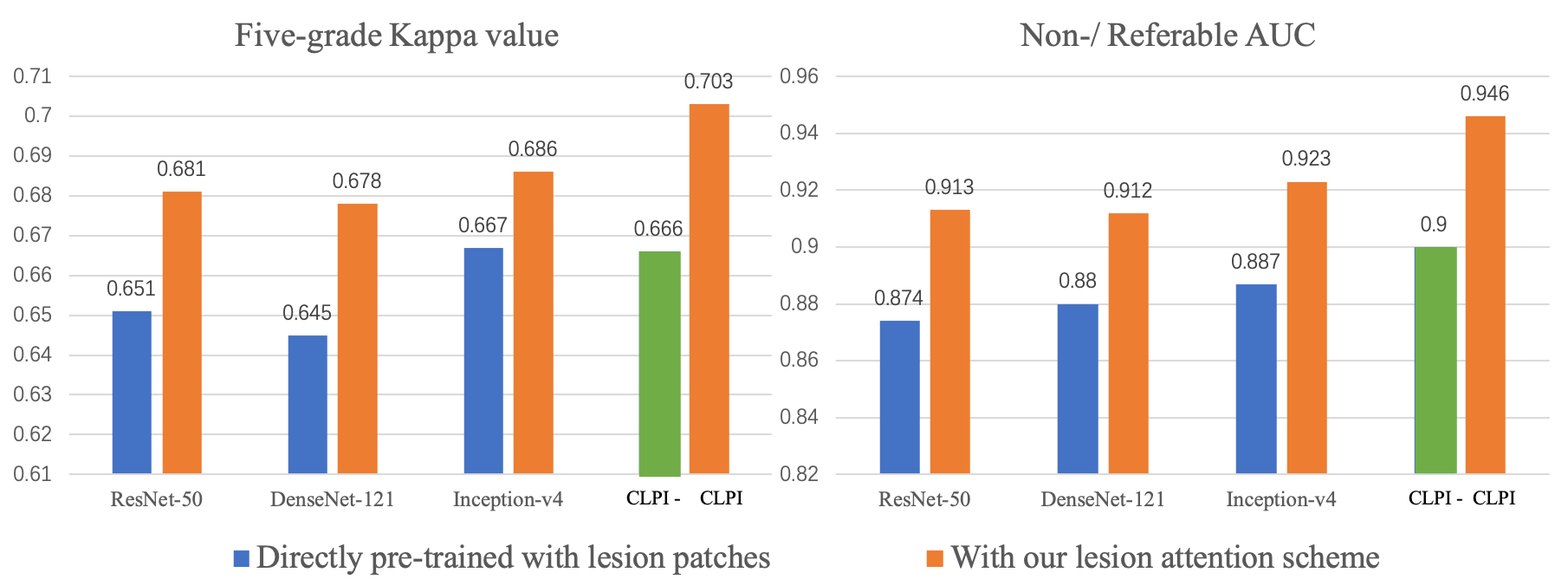}
\caption{Both the five-grade Kappa value and AUC of binary DR grading are improved by using our lesion attention generator. The blue bars indicate the classifiers pre-trained with the lesion patches directly. The orange bars show the results of the same methods that trained with lesion attention maps provided by our lesion attention generator. The green bars present the results of CLPI without patch pretraining, named as CLPI-.}
\label{fig.lesionAttentionAdvantage}       
\end{figure}

\noindent {\bf The effectiveness of the lesion attention generator.} 
To prove that our lesion attention generator is helpful for DR grading, we train some popular classification architectures directly with our weighted attention map $\bf{I^{AM}}$ (see Fig.~\ref{fig.mainWorkflow}) instead of the fundus images. The $\bf{I^{AM}}$ is achieved offline by our lesion attention generator. Fig.~\ref{fig.lesionAttentionAdvantage} illustrates the results of the classification architectures trained with the private dataset and tested on {\em Messidor-2} dataset. 
All the methods are pre-trained with the patches from {\em IDRiD} dataset except CLPI-.  CLPI- is the proposed framework without patch-level pre-training. We can see that each method trained with our lesion attention generator outperforms the one directly pre-trained with lesion patches. 
Therefore, our lesion attention scheme is effective in improving DR grading. In addition, the comparison between CLPI and CLPI- shows the merits of exploring lesion features for DR grading.

Additionally, the five-grade confusion matrices of CLPI and the DenseNet-121 are listed in Table~\ref{table:confmat1}.
We can see that most errors take place around the adjacent grades. Such mistakes are not easy to avoid even by clinical ophthalmologist. It is unacceptable that the severe NPDR (Label 3) / PDR (Lable 4) be miss-classified into No DR (Label 0) or mild DR (Label 1) and vice versa. The bottom left (red) and top right (blue) of the confusion matrices in Table~\ref{table:confmat1} record the unacceptable errors. Compared to the Densenet-121, the total number of the unacceptable errors is reduced from $34$ to $20$.  Note that the grading module of CLPI and DenseNet-121 share the same backbone, the reduction of the unacceptable mistakes further demonstrates the effectiveness of our lesion attention generator.

\begin{figure}[!htb]
\centering
\includegraphics[width=3.3in]{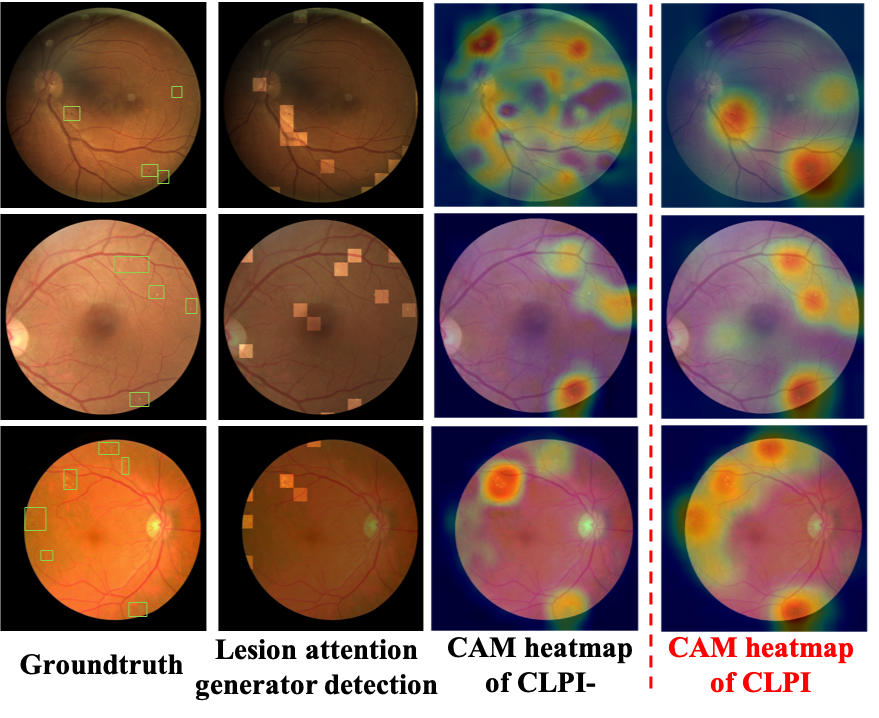}
\caption{The qualitative comparison of abnormal region locating. By collaboratively integrating lesion and grade annotations, CLPI can focus on the lesion regions more accurately.}
\label{fig.camCmp}       
\end{figure}

\begin{table}
\caption{The confusion matrices of five-grade evaluation. The models are trained and testing with {\em Private} dataset under the same circumstances}
\centering
\begin{tabular}{ccccccc}
\hline
   &   & \multicolumn{5}{c}{Prediction} \\
 Method                     & Label                  & 0     & 1   & 2    & 3   & 4   \\ \hline
     \multirow{5}{*}{DenseNet-121}  & \multicolumn{1}{c|}{0} & 6188  & 260 & 145  & \textcolor{blue}{0}   & \textcolor{blue}{16}  \\
  & \multicolumn{1}{c|}{1} & 220   & 653 & 201  & \textcolor{blue}{1}   & \textcolor{blue}{1}   \\
  & \multicolumn{1}{c|}{2} & 145   & 311 & 2580 & 173 & 40  \\
   & \multicolumn{1}{c|}{3} & \textcolor{red}{4}     & \textcolor{red}{0}   & 286  & 378 & 24  \\
  & \multicolumn{1}{c|}{4} & \textcolor{red}{12}    & \textcolor{red}{0}   & 93   & 58  & 307 \\ \cline{1-7} 
\multirow{5}{*}{CLPI}         & \multicolumn{1}{c|}{0} & 6258  & 212 & 128  & \textcolor{blue}{7}   & \textcolor{blue}{4}   \\
  & \multicolumn{1}{c|}{1} & 218   & 655 & 202  & \textcolor{blue}{0}   & \textcolor{blue}{1}   \\
  & \multicolumn{1}{c|}{2} & 139   & 207 & 2745 & 127 & 31  \\
  & \multicolumn{1}{c|}{3} & \textcolor{red}{3}     & \textcolor{red}{0}   & 288  & 368 & 33  \\
  & \multicolumn{1}{c|}{4} & \textcolor{red}{3}     & \textcolor{red}{2}   & 65   & 45  & 355 \\ \hline
\end{tabular}
\label{table:confmat1}
\end{table}

\noindent {\bf The necessity of the end-to-end collaborative learning scheme.}
In this part, we set up ablation studies from both quantitative and qualitative aspects.  

Table \ref{tab.CL-shortcut-cmp} records the quantitative ablation studies. Wherein, {\em CLPI without end-to-end learning} means that training the CLPI by fixing the pre-trained lesion attention generator. In this way, the information is only one-way transmit from the lesion attention generator to the grading module, therefore the lesion attention generator cannot be optimized by the image-level grade annotations.
{\em CLPI without shortcut} denotes that training CLPI framework without the lesion shortcut in Fig.~\ref{fig.mainWorkflow}, i.e., the gradient are back-propagated through the {\bf deep} classification backbone to the lesion attention generator.

As seen in Table~\ref{tab.CL-shortcut-cmp}, the outperformance of CLPI over {\em CLPI without end-to-end learning} quantitatively reveals that end-to-end learning is necessary to fully integrate the lesion and grade information. In addition, the comparison between CLPI and {\em CLPI without shortcut} indicates the motivation of introducing the shortcut is effective and reasonable.

Fig. \ref{fig.camCmp} illustrates qualitative comparisons in an interpretable way, in which the first column shows the groundtruth of our lesion annotation in terms of MA, hemorrhage and exudate. The second column records the detection results of our lesion attention generator, and the highlighted boundingboxes are predictions that contain at least one of the three lesions. The third and fourth columns are the class activation map (CAM) \cite{CAM2016} heatmaps of CLPI- (without lesion pre-training) and CLPI respectively, and the redder regions indicate the higher abnormal probabilities. CAM is an interpretable way to visualize the class-related heatmaps in the CNNs-based classifiers.  All the CAMs are extracted from the last convolution layer of the networks, and readers can refer to \cite{CAM2016} for more details. 

As seen in Fig.~\ref{fig.camCmp}, the CAMs of CLPI focus on the lesion regions more accurately than the lesion detection results as well as CLPI-. This reveals that 1) the lesion information from the lesion attention generator can be refined by the image-level grade information for CLPI; 2) The {\em black box} DR severity classifier via image-level grade annotations is relatively weak in lesion-related feature learning. Therefore, the CLPI framework can build information exchange between lesion and grade annotations. 

Additionally, we can see most lesion regions are caught by both the CAMs of grading net and CLPI in Fig.~\ref{fig.camCmp}, which reveals that the lesions have high-relevance with the final decision of the CNNs-based grading networks. This study proves the correlation between the lesion and grade in an experimental aspect, which confirms the reasonability of exploiting lesion information for automatically DR grading. 

To sum up both the quantitative and qualitative ablation study: by end-to-end optimizing the final framework to collaboratively build bidirectional information exchange, CLPI achieves more discriminative features to improve the performance of DR grading.

\begin{table*}[!htb]
\caption {Ablation studies w.r.t. shortcut and collaborative learning scheme. The table lists the Kappa values of DR grading.}
\centering
\footnotesize
\label{tab.CL-shortcut-cmp}
\begin{tabular}{l|cc|cc|cc}
\hline
Testing sets  & \multicolumn{2}{c|}{\em Private} & \multicolumn{2}{c|}{\em Messidor-2} & \multicolumn{2}{c}{\em LIQ-EyePACS}\\
\hline
Methods & \begin{tabular}[c]{@{}c@{}}Five-grade\\ (Kappa)\end{tabular} & \begin{tabular}[c]{@{}c@{}}Binary\\ (AUC)\end{tabular} & \begin{tabular}[c]{@{}c@{}}Five-grade\\ (Kappa)\end{tabular} & \begin{tabular}[c]{@{}c@{}}Binary\\ (AUC)\end{tabular} & \begin{tabular}[c]{@{}c@{}}Five-grade\\ (Kappa)\end{tabular} & \begin{tabular}[c]{@{}c@{}}Binary\\ (AUC)\end{tabular}\\
\hline
\begin{tabular}[l]{@{}l@{}}{\em CLPI without} \\{\em end-to-end learning}\end{tabular}   & 0.905  & 0.971 & 0.674  & 0.909 & 0.647 & 0.903  \\
\hline
{\em CLPI without shortcut} & 0.903  & 0.973 & 0.675  & 0.933 & 0.665 & 0.911\\
\hline
\textbf{CLPI} & \textbf{0.908} & \textbf{0.983} & \textbf{0.703} & \textbf{0.946} & \textbf{0.692} & \textbf{0.916} \\
\hline
\end{tabular}
\vspace{-0.5cm}
\end{table*}

\section{Discussion} \label{sec.discussion}

The above experiments and ablation studies have proved the effectiveness and robustness of the proposed algorithm. By introducing a few patch-level lesion annotations, the proposed framework can collaboratively integrate both lesion and image-level grade information to achieve promising results. The proposed lesion attention generator and shortcut connection are carefully designed to facilitate the end-to-end training of the framework. 

\begin{figure}[!htb]
\centering
\includegraphics[width=3.3in]{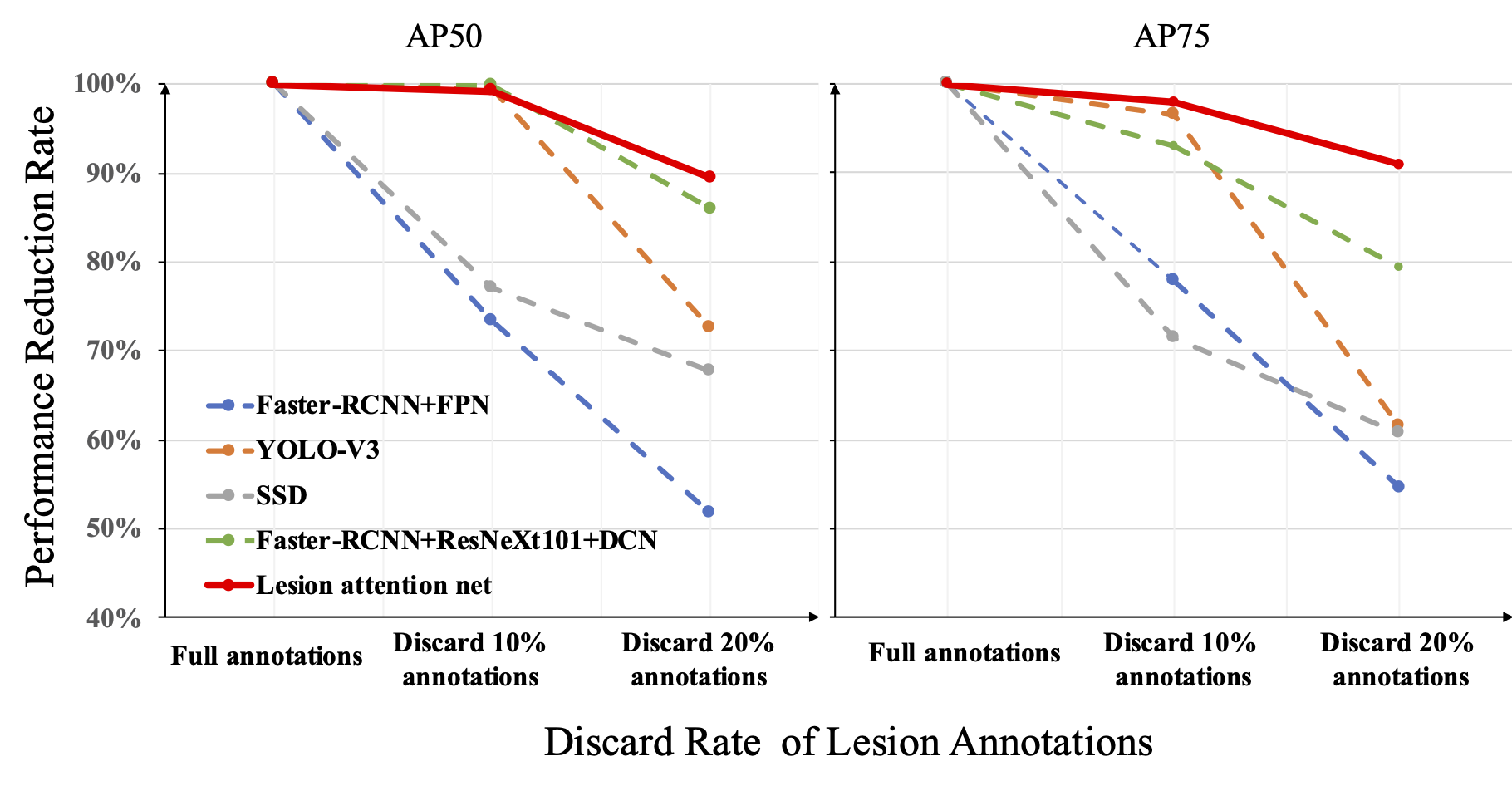}
\caption{The performance reduction rate corresponding to the discard rate of lesion annotations. a) and b) respectively record the comparisons of AP50 and AP75 metrics in lesion detection. }
\label{fig.lesionRobust}       
\end{figure}

Additionally, when our lesion attention generator is served as a lesion detector, we find that it is more robust to missing labels compared to some SOTA detectors including Fast-RCNN \cite{FasterRCNN}, YOLO-V3 \cite{yoloV3}, SSD \cite{ssd} and Faster-RCNN + ResNeXt10 backbone + deformable convolution network \cite{DCN2017}. In Fig.~\ref{fig.lesionRobust}, we randomly discard some lesion annotations in the training dataset, and record the performance reduction rates ($\frac{performance\_after\_discard}{performance\_with\_full\_annotations}$). In this case, the discarded annotations will turn to missing-annotated samples for the detectors trained with entire images, but the negative patches for training our lesion detection net can be collected from the No-DR images to avoid the missing labels. Therefore, the SOTA detectors are sensitive to the lack of the annotations while the proposed detection architecture shows relatively robust performance.

Although the proposed lesion attention generator can effectively exploit lesion information and the architecture is robust to missing labels, it can only serve as an attention generator rather than a lesion detector. This is because the proposed architecture does not take scale into consideration when generate proposals, and it does not have a regression module to get precise boundingboxes. Besides, we only cover three key lesions among a dozen of DR-related lesions.
However, with limited lesion annotations, our lesion attention generator is elegant and qualified to achieve valuable attention for improving the performance of DR grading. 

CLPI and Semi + Adv \cite{zhouCVPR2019} share similar motivations to collaboratively integrate the lesion and grade information. However, the two methods are totally different in architecture designing. As a result, CLPI achieves comparable performance compared to \cite{zhouCVPR2019}, while only need patch-level lesion annotations instead of pixel-level masks used by \cite{zhouCVPR2019}. It is worth noting that the patch-level annotations are much easier to be achieved as well as time and labor-saving compared to pixel-level annotations. 

Moreover, the proposed framework can also be extended to other classification tasks that the objects in the image are related to the image category, such as scene classification and other disease grading tasks based on medical images. We have not evaluated our framework on these tasks in this paper, and it's our future work to make CLPI more general. 

\section{Conclusion}

In this paper, we propose a collaboratively learning framework to integrate patch-level lesion and image-level grade features for robust DR grading. 
The lesion attention generator provides valuable semi-supervised lesion attention for DR grading, while the grade supervision is back-propagated to optimize the attention map.
Our lesion attention generator is trained with patches, yet it can detect lesions over an entire image in only one forward pass. This
facilitates the end-to-end learning of the entire framework.
Extensive experiments prove that the proposed CLPI has comparable performance with SOTA algorithms as well as senior ophthalmologists. We show the robustness of CLPI by evaluating the DR grading methods under various challenges in real-world scenarios. Ablation studies have shown the effectiveness of the lesion attention scheme as well as the advantages of the end-to-end collaborative learning of CLPI. Some open issues still exist, including precisely detecting more types of lesions and extending CLPI framework to other applications besides DR grading.

%

\appendices

\section{The details of the lesion attention map generation}\label{appendix.attentionMapGen}

\begin{figure}
\centering
\includegraphics[width=3.3in]{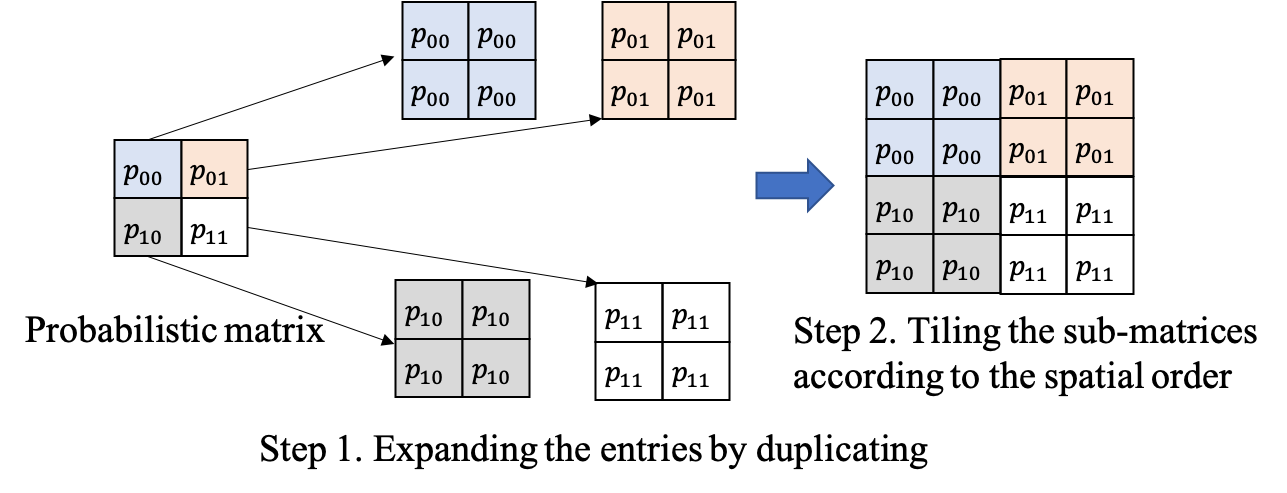}
\caption{Illustration of expanding a $2 \times 2$ probabilistic matrix to a $4 \times 4$ attention map.}
\label{fig:expanding}       
\end{figure}

There are two steps to turn the probabilistic matrix $\mathbf{P}_i$ into attention map: 1) expending each entry of $\mathbf{P}_i$ to a sub-matrix by duplicating; and 2) tiling all the sub-matrices generated by $\mathbf{P}_i$ according to the spatial order. Figure~\ref{fig:expanding} illustrates how a $2 \times 2$ matrix be expanded to a $4 \times 4$ with our scheme. In our paper,  $\mathbf{P}_i \in \mathbb{R}^{16 \times 16}$ will be turned to a $512 \times 512$ attention map. Therefore, every entry of $\mathbf{P}_i$ is expanded to a $32 \times 32$ ($32$ is achieved by $512 / 16$) sub-matrix by duplicating. Then all the $16 \times 16$ sub-matrices are tiled into $512 \times 512$ matrix according to the spatial orders in $\mathbf{P}_i$.

\section*{Acknowledgment}
We would thank the Post-Doctoral Workstations of Baidu and Beijing Institute of Technology. These workstations have provide plenty of support for this research.


\ifCLASSOPTIONcaptionsoff
  \newpage
\fi



%

\bibliographystyle{IEEEtran}
\bibliography{main}{}




%






\end{document}